 \documentclass[pmlr,twocolumn,10pt]{jmlr} % W&CP article

\usepackage{booktabs}
\usepackage{siunitx}
\usepackage{stfloats,multicol,lipsum}

\newcommand{\equal}[1]{{\hypersetup{linkcolor=black}\thanks{#1}}}

\theorembodyfont{\upshape}
\theoremheaderfont{\scshape}
\theorempostheader{:}
\theoremsep{\newline}

\jmlrvolume{LEAVE UNSET}
\jmlryear{2023}
\jmlrsubmitted{LEAVE UNSET}
\jmlrpublished{LEAVE UNSET}
\jmlrworkshop{ } % W&CP title

\title[HeAR - Health Acoustic Representations]{HeAR - Health Acoustic Representations}

\author{
\Name{Sebastien Baur$^1$}\equal{These authors contributed equally}  
\Name{,Zaid Nabulsi$^1$}\footnotemark[1] 
\Name{,Wei-Hung Weng$^1$,} 
\Name{Jake Garrison$^1$,} 
\Name{Louis Blankemeier$^1$,}
\Name{Sam Fishman$^1$,}
\Name{Christina Chen$^1$,}
\Name{Sujay Kakarmath$^1$,}
\Name{Minyoi Maimbolwa$^2$,}
\Name{Nsala Sanjase$^2$,}
\Name{Brian Shuma$^2$,}
\Name{Yossi Matias$^1$,}
\Name{Greg S. Corrado$^1$,}
\Name{Shwetak Patel$^1$,}
\Name{Shravya Shetty$^1$,}
\Name{Shruthi Prabhakara$^1$,}
\Name{Monde Muyoyeta$^2$,}
\Name{Diego Ardila$^1$} \\
\addr $^1$Google Research, USA\\
\addr $^2$TB department, Center of Infectious Disease Research in Zambia, Zambia\\
\Email{\url{health\_acoustic\_representations@google.com}}
}
\begin{document}
\onecolumn

\maketitle

\begin{abstract}
Health acoustic sounds such as coughs and breaths are known to contain useful health signals with significant potential for monitoring health and disease, yet are underexplored in the medical machine learning community. The existing deep learning systems for health acoustics are often narrowly trained and evaluated on a single task, which is limited by data and may hinder generalization to other tasks. To mitigate these gaps, we develop HeAR, a scalable self-supervised learning-based deep learning system using masked autoencoders trained on a large dataset of 313 million two-second long audio clips. Through linear probes, we establish HeAR as a state-of-the-art health audio embedding model on a benchmark of 33 health acoustic tasks across 6 datasets. By introducing this work, we hope to enable and accelerate further health acoustics research. 
\end{abstract}

% \vspace{-15pt}
\section{Introduction}
% \vspace{-5pt}
\label{sec:intro}
Acoustic non-semantic attributes of speech can enable machine learning models to perform paralinguistic tasks, including emotion recognition, speaker identification, and dementia detection~\citep{shor2022universal}. Cerebrovascular and neurodegenerative diseases like stroke, Parkinson's, Alzheimer's, cerebral palsy and amyotrophic lateral sclerosis (ALS) may also be detected and monitored using non-semantic patterns of speech, such as articulation, resonation, and phonation~\citep{boschi2017connected}. Non-semantic acoustic signals related to health are not confined solely to conversational speech data. Health-related acoustic cues, originating from the respiratory system's airflow, including sounds like coughs and breathing patterns can be harnessed for health monitoring purposes. For example, clinicians use sounds such as ``whoop''-like coughing to diagnose pertussis~\citep{pramono2016cough}, and agonal breathing for detecting acute cardiovascular events. Such health sounds can also be collected via ambient sensing technologies on ubiquitous devices such as mobile phones~\citep{zimmer2022making}, which may augment healthcare workers in low-medium income countries (LMICs) with improved screening capabilities.

With advancements in deep learning, neural networks are now able to learn high-quality general representations directly from raw speech data~\citep{zhang2022bigssl}, and use them for various semantic and non-semantic speech-related tasks~\citep{peplinski2020frill,shor2022universal,shor2022trillsson}. Health acoustics, specifically non-semantic respiratory sounds, also have potential as biomarkers to detect various respiratory diseases~\citep{alqudaihi2021cough}. However, current machine learning (ML) systems for health acoustics are task-specific and may not generalize well to out-of-distribution (OOD) settings~\citep{d2022underspecification} and are often limited by data quantity.

Recently, self-supervised learning (SSL) has demonstrated potential for building robust and capable systems by learning general representations from large, unlabeled sources~\citep{balestriero2023cookbook,chen2020simple,he2022masked}. There is extensive progress on learning general, universal representations in vision~\citep{dosovitskiy2020image}, language~\citep{chowdhery2022palm} and speech~\citep{zhang2023google}. Such approaches are also used for learning representations of biomedical language, medical images and even physiological waveforms~\citep{belyaeva2023multimodal,singhal2023large,xu2023elixr}. However, these approaches are still underexplored in the field of health acoustics.

To demonstrate the potential of SSL on the underexplored health acoustic modality, we introduce \textbf{HeAR: Health Acoustic Representations}, a self-supervised generative learning-based system trained on a large dataset of two-second long audio clips for learning low-dimensional representations that can transfer well across health acoustic tasks and generalize to OOD data. We benchmark HeAR on a diverse set of health acoustic tasks spanning 13 health acoustic event detection tasks, 14 cough inference tasks, and 6 spirometry inference tasks, across 6 datasets and demonstrate that simple linear classifiers trained on top of our representations outperform the state-of-the-art on many tasks.

% \vspace{-15pt}
\section{Related Works}
% \vspace{-5pt}
\label{sec:related}
SSL has emerged as a critical ML paradigm to learn general representations from large unannotated datasets. SSL can have numerous training objectives including contrastive, such as SimCLR~\citep{chen2020simple}, BYOL~\citep{grill2020bootstrap}, and generative, like masked autoencoder (MAE)~\citep{he2022masked}. In recent years, data-driven audio SSL has made great progress, specifically for semantic speech. From CPC~\citep{oord2018representation}, Wav2vec 2.0~\citep{baevski2020wav2vec}, BigSSL~\citep{zhang2022bigssl}, AudioMAE~\citep{huang2022masked}, to BEST-RQ~\citep{chiu2022self} and Universal Speech Model (USM)~\citep{zhang2023google}, researchers utilize massive unlabeled data from the Internet to train SSL-based audio encoders to learn better speech representations. There are also studies focusing on non-semantic speech. For example, TRILL~\citep{shor2020towards} uses a triplet loss as the training objective, TRILLsson~\citep{shor2022trillsson} and FRILL~\citep{peplinski2020frill} further distill the TRILL encoder to make it smaller and faster.  Researchers also adopted different neural network architectures, such as Conformer~\citep{shor2022universal,srivastava2022conformer}, and Slowfast NFNet~\citep{kazakos2021slow,wang2022towards} to develop performant audio encoders. In our work, we adopt a generative SSL framework (MAE) but focus on non-semantic health acoustics, which is relatively underexplored yet useful for healthcare applications.

ML for non-semantic health acoustics is also an emerging research area for health. There are a growing number of studies using respiratory sounds for health monitoring and disease detection. For example, cough sound patterns can be used as a biomarker to identify coughers~\citep{whitehill2020whosecough}, detect various respiratory diseases, such as COVID-19~\citep{coppock2021end,laguarta2020covid,schuller2020detecting}, bronchitis, bronchiolitis, pertussis~\citep{bales2020can}, obstructive versus restrictive lung diseases~\citep{rudraraju2020cough}, and tuberculosis (TB)~\citep{tracey2011cough,larson2012validation,botha2018detection,rudraraju2020cough,pahar2021automatic,zimmer2022making,sharma2024tbscreen}. However, these works focus on developing a single task-specific system trained in a supervised learning framework that may not generalize as well to new settings. Our work instead introduces a system trained without supervision on a large and diverse unlabelled corpus, which may generalize better to unseen distributions and new tasks~\citep{radford2021learning}.

% \vspace{-15pt}
\section{Methods}
\begin{figure*}[h]
\centering
\includegraphics[width=0.9\linewidth]{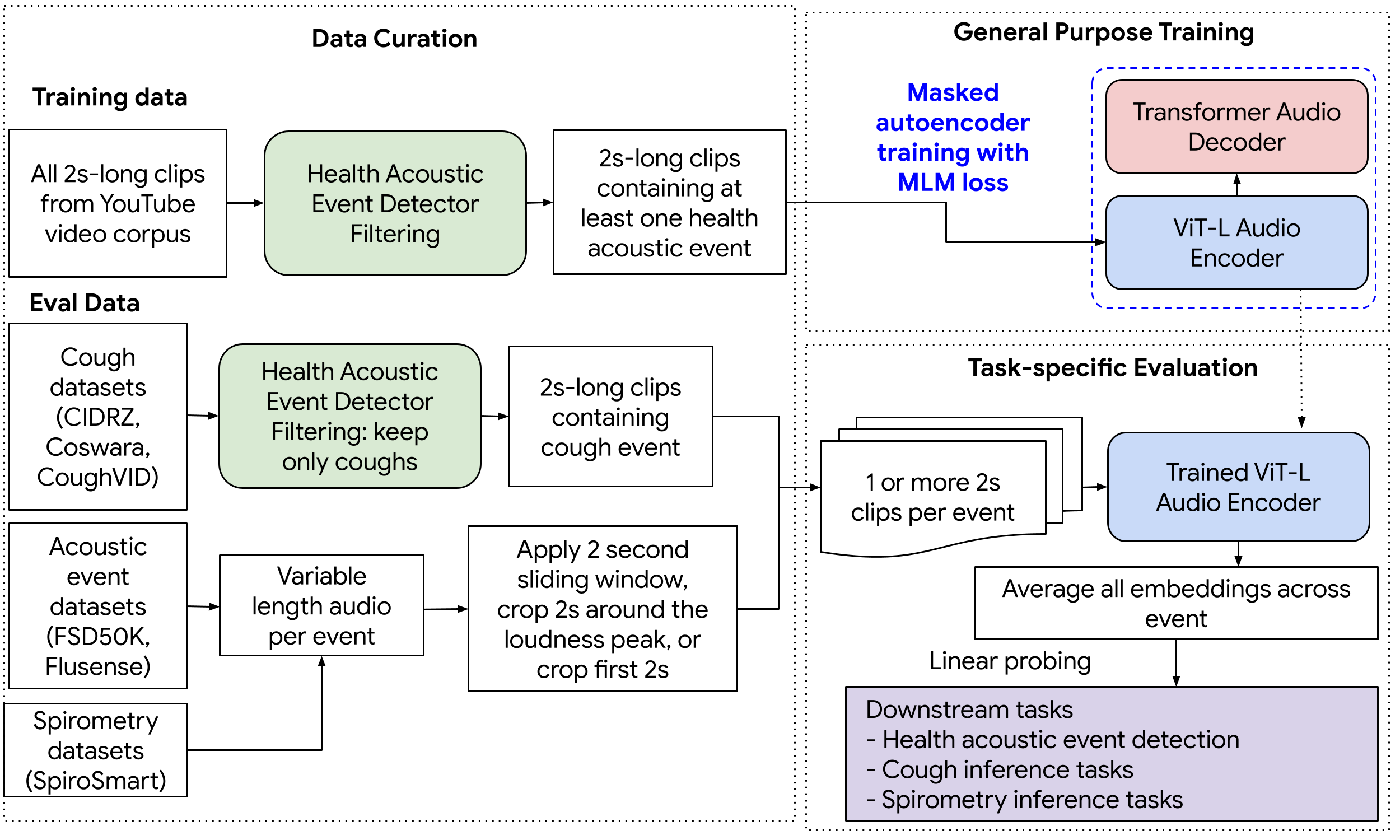} 
\caption{HeAR system overview.}
\label{fig1}
\end{figure*}

% \vspace{-5pt}
\label{sec:methods}
HeAR comprises three main components: a data curation step (including a health acoustic event detector), a general purpose training step to develop an audio encoder (embedding model), and a task-specific evaluation step that adopts the trained embedding model for various downstream tasks. The system is designed to encode two-second long audio clips and generate audio embeddings for use in downstream tasks. Figure~\ref{fig1} illustrates these high-level components of the system.

The health acoustic event detector is a multilabel classification convolutional neural network (CNN) that identifies the presence of any of six types of non-speech health acoustic events in two-second audio clips : coughing, baby coughing, breathing, throat clearing, laughing, and speaking. The detector is described in Appendix A.

Similarly to AudioMAE~\citep{huang2022masked}, we used a MAE~\citep{he2022masked} to learn audio representations by training an autoencoder to reconstruct masked 16x16 spectrogram patches (units: time and frequency). 75\% of input patches are masked out and encoded by ViT-L~\citep{dosovitskiy2020image}. Learnable mask tokens are added to the sequence of encoded tokens, and an 8-layer transformer decoder is tasked with reconstructing the missing patches, by minimizing the $L_2$ distance between normalized masked patches and its predictions. Note that we did not experiment with self-attention as done by~\cite{huang2022masked}. The model is trained using the AdamW optimizer for 950k steps ($\sim$4 epochs) with global batch size 4096, and hyperparameters from~\cite{huang2022masked}. Learning rate follows a cosine decay schedule, starting at 4.8e-4, following the commonly used linear batch scaling rule~\citep{goyal2017accurate}.

% \vspace{-5pt}
\paragraph{Datasets}
For training, we curate a dataset, YT-NS (YouTube Non-Semantic), consisting of two-second long audio clips extracted from three billions public non-copyrighted YouTube videos using the health acoustic event detector (described in Appendix A), totalling 313.3 million two-second clips or roughly 174k hours of audio. We chose a two-second window since most events we cared about were shorter than that. The HeAR audio encoder is trained solely on this dataset.

We benchmark HeAR both on general health acoustic event classification and on cough inference tasks. For general health acoustic event classification, we use FSD50K~\citep{fonseca2021fsd50k} and FluSense~\citep{al2020flusense}. FluSense contains human-annotated timestamped labels, which we use to extract short labeled audio clips (50\% are shorter than 2 seconds and 90\% shorter than 7 seconds). FSD50K does not have such timestamps, and some of its clips can be far longer than the duration of clips seen during training. For this reason, we crop two-second audio clips around the loudness peak, computed as the average sound amplitude in dB. We found that this resulted in higher performance for all models on all detection tasks. 

For cough inference tasks, we use~\href{https://zenodo.org/record/4498364#.Yf21afXMJqv}{CoughVID}~\citep{orlandic2021coughvid},~\href{https://github.com/iiscleap/Coswara-Data}{Coswara}~\citep{bhattacharya2023coswara}, and a prospective tuberculosis-specific dataset from the Centre for Infectious Disease Research in Zambia (CIDRZ) where cough audio recordings and chest X-rays were obtained from a cohort of symptomatic patients. Considering that the recording quality may be affected by various environmental factors, we obtain CIDRZ audio from microphones of varying quality under a study data collection protocol that controls the environmental factors. Three devices were used to collect the cough audio, representing three mobile phone ``tiers'' (for the purposes of this study) with different costs and potentially varying recording quality: Pixel3a, GalaxyA12, and GalaxyA22. We refer to the datasets as CIDRZ low-tier, CIDRZ mid-tier, and CIDRZ high-tier respectively. We use recordings from all devices to train the linear probes since they provide more data, and focus evaluation results in the main text on one device (low-tier) for brevity. Results for evaluation on mid-tier and high-tier datasets are available in Appendix C. We also include evaluation on SpiroSmart, one pulmonary testing dataset described in~\cite{garrison2018spiro}, which includes spirometry efforts paired with audio recordings of forced expiratory efforts from patients of chronic obstructive pulmonary disease (COPD) clinics around the world. We describe the evaluation datasets, including train/validation/test split sizes, in Appendix Table~\ref{tab_b1}, and summary statistics of the unpublished CIDRZ dataset are listed in Appendix B and Appendix Table~\ref{tab_b2}.

% \vspace{-5pt}
\paragraph{Baseline Models}
We consider several state-of-the-art audio encoder baselines for comparison: (1) TRILL~\citep{shor2020towards}, a publicly available ResNet50-based encoder trained on an AudioSet subset with speech labels by optimizing triplet loss, (2) FRILL~\citep{peplinski2020frill}, a publicly available MobileNet-based distilled version of TRILL for mobile devices, (3) BigSSL-CAP12~\citep{shor2022universal}, a Conformer-based encoder trained on YouTube 900k-hour speech and LibriLight with a wav2vec 2.0 objective, and (4) CLAP, a CNN-based audio encoder trained using multimodal contrastive learning on a mixture of datasets that include FSD50K~\citep{elizalde2023clap}. We also investigated different Mel spectrogram-based models using spectrograms or Mel-frequency cepstral coefficients (MFCCs) as features, as well as a randomly initialized MAE. We didn't include them in the final evaluation since their performance was very low and they did not contribute meaningfully to our benchmark.

% \vspace{-5pt}
\paragraph{Evaluation on Downstream Tasks}
To evaluate the quality of representations learned by HeAR and compare it to other encoders', we train linear probes. More specifically, we encode all recordings from all datasets using TRILL, FRILL, BigSSL-12, HeAR, and CLAP, and we train separate linear or logistic regressions (with a cross-validated ridge penalty) to predict available labels on these datasets~\citep{kohn2015s}. When this information is available, the cross-validation procedure groups recordings of the same individual, the same audio clip, and the same site within the same fold, to stick as close as possible to the out-of-sample evaluation scenario. Once the regularization coefficient is chosen, we evaluate the performance of that linear model on a held-out validation dataset for which all audio recordings come from sources not seen for training HeAR or cross-validating the linear probes. The validation datasets were used to experiment with various data preprocessing schemes, especially for datasets that had clips longer than YT-NS (up to 30x longer). Finally, once the best preprocessing scheme has been identified on the validation datasets, we compute the performance of the best linear models on our held-out test dataset, which is also disjoint from the train and validation datasets. We set up 13 health acoustic event detection tasks from two datasets, 14 cough inference tasks from three datasets, and 6 spirometry tasks from one dataset. For FluSense and FSD50K, we train a separate linear probe for each task, predicting whether the specific audio event occurs in the clip. For the cough inference tasks, we train a separate linear probe for each task and for each dataset, predicting the specific label from a two-second audio recording of a cough. Cough inference tasks include identifying three types of chest X-ray (CXR) findings (unspecified abnormalities, presence of focal or multifocal lung opacities, and pleural effusion), two diagnostic tasks (COVID on two datasets and tuberculosis on another one), and identifying demographics and lifestyle factors (smoking status, sex, age, BMI). Spirometry tasks include estimation of forced expiratory volume (FEV1), forced vital capacity (FVC), the FEV1/FVC ratio, peak flow (PEF), total exhale duration (FET), and sex classification. We report either the area under the receiver operating characteristic curve (AUROC) or average precision (AP) for one-versus-rest classification tasks. We used the DeLong method to compute the 95\% confidence intervals (CIs) of AUROC~\citep{delong1988comparing}. For regression tasks, we report mean absolute error with bootstrapped 95\% confidence intervals.

% \vspace{-15pt}
\section{Results}
\begin{figure}[h]
\centering
\includegraphics[width=0.7\linewidth]{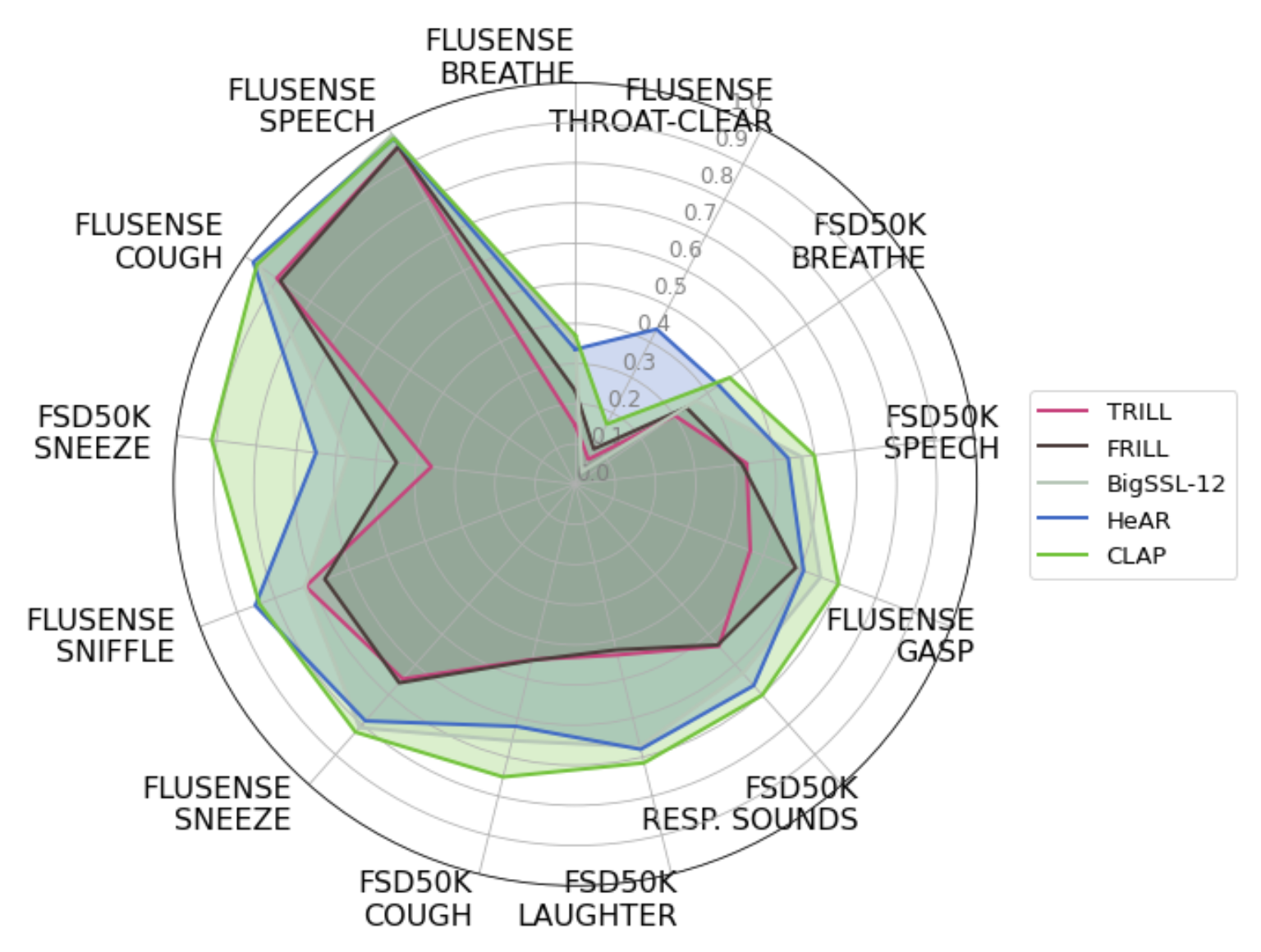} 
\caption{Radar plot of the performance comparison on health acoustic event detection tasks on FSD50K and FluSense datasets.}
\label{fig2}
\end{figure}

\begin{figure}[h]
\centering
\includegraphics[width=0.7\linewidth]{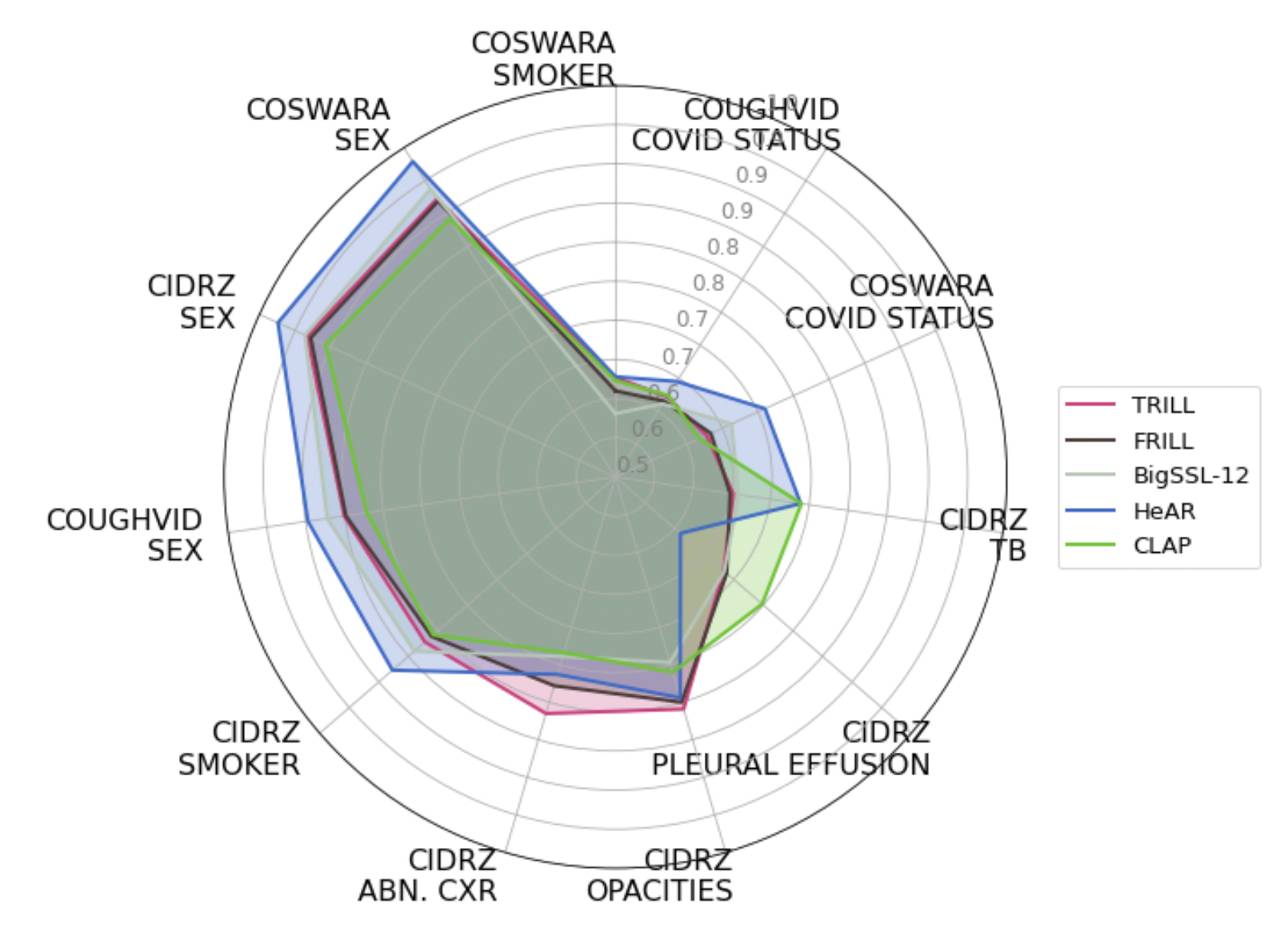} 
\caption{Radar plot of the performance comparison on cough inference tasks.}
\label{fig3}
\end{figure}

\begin{figure}[h]
\centering
\includegraphics[width=0.7\linewidth]{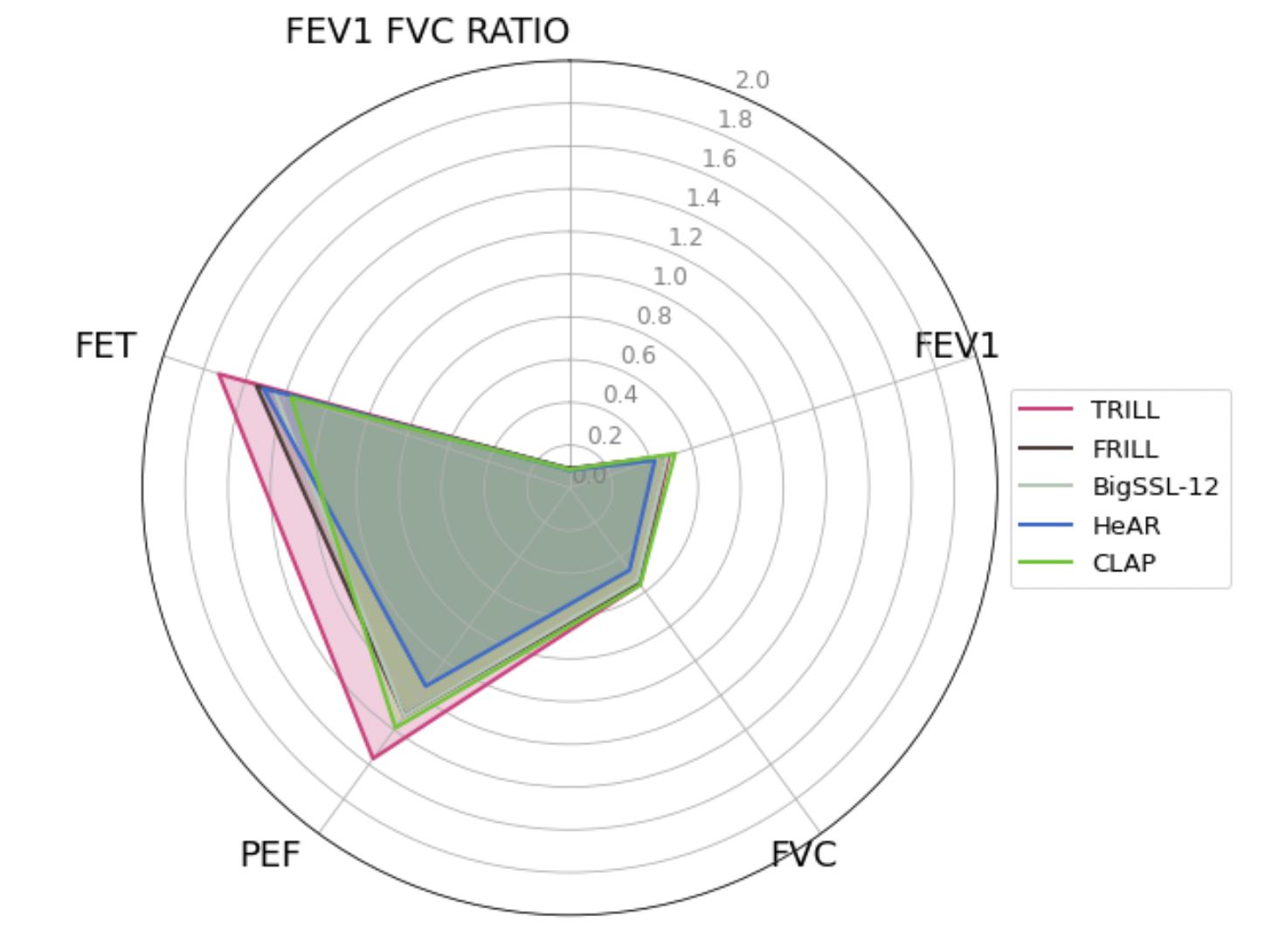} 
\caption{Radar plot of the performance comparison on spirometry tasks.}
\label{fig4}
\end{figure}

% \vspace{-5pt}
Across a range of 33 tasks on 6 datasets, HeAR achieved the highest performance among all models, as measured by mean reciprocal rank (0.708, see Table~\ref{tab1}), reaching the highest rank on 17 tasks (3 out of 13 health acoustic event detection tasks in Table~\ref{tab2} and Figure~\ref{fig2}, 10 out of 14 cough inference tasks in Table~\ref{tab3} and Figure~\ref{fig3}, and 5 out of 6 spirometry tasks in Table~\ref{tab4} and Figure~\ref{fig4}).

\begin{table*}[h]
\footnotesize
\centering
\begin{tabular}{cccccc}
\toprule
\textbf{Task group} & \textbf{TRILL} & \textbf{FRILL} & \textbf{BigSSL-CAP12} & \textbf{HeAR} & \textbf{CLAP (48k)} \\
\midrule
All & 0.322 & 0.273 & 0.419 & \textbf{0.708} & 0.555 \\
Health acoustic detection & 0.225 & 0.235 & 0.438 & 0.538 & \textbf{0.846} \\
Cough & 0.423 & 0.305 & 0.373 & \textbf{0.812} & 0.370 \\
Spirometry & 0.298 & 0.281 & 0.486 & \textbf{0.833} & 0.356 \\
\bottomrule
\end{tabular}
\caption{Mean reciprocal ranks on groups of tasks.}
\label{tab1}
\end{table*}

For the health acoustic detection tasks, CLAP performs best overall (mAP=0.691, MRR=0.846), which might be expected since FSD50K was used in its training procedure. HeAR has the second highest performance (mAP=0.658, MRR=0.538) and highest among models that haven't used FSD50K for training. We observed on FSD50K that the performance of HeAR degraded significantly with sequence length, which we hypothesize to be due to the use of fixed sinusoidal positional encodings, which are known~\citep{kazemnejad2024impact} to generalize poorly to unseen longer sequence lengths. Cropping the loudest two-second clip improved the performance of all models, and especially HeAR. 

\begin{table*}[h]
\centering
\resizebox{\textwidth}{!}{
\begin{tabular}{ccccccccc}
\toprule
\textbf{Dataset} & \textbf{\begin{tabular}[c]{@{}c@{}}Binary\\ Classification\\ Task\end{tabular}} & \textbf{\begin{tabular}[c]{@{}c@{}}Summary statistics\\ for test split\\ (label)\end{tabular}} & \textbf{Metric} & \textbf{TRILL} & \textbf{FRILL} & \textbf{BigSSL-CAP12} & \textbf{HeAR} & \textbf{CLAP (48k)} \\
\midrule
\begin{tabular}[c]{@{}c@{}}FSD50K + \\ FluSense\end{tabular} & All & N/A & mAP & 0.494 & 0.516 & 0.613 & 0.658 & \textbf{0.691} \\
\midrule
FSD50K & Breathing & 227 / 10004 (Y/N) & AP & 0.301 {[}0.242, 0.365{]} & 0.336 {[}0.276, 0.399{]} & 0.365 {[}0.294, 0.434{]} & 0.434 {[}0.365, 0.496{]} & \textbf{0.467 {[}0.394, 0.538{]}} \\
 & Cough & 106 / 10125 (Y/N) & AP & 0.450 {[}0.356, 0.547{]} & 0.452 {[}0.359, 0.543{]} & 0.658 {[}0.568, 0.742{]} & 0.621 {[}0.513, 0.719{]} & \textbf{0.751 {[}0.673, 0.821{]}} \\
 & Laughter & 253 / 9978 (Y/N) & AP & 0.438 {[}0.379, 0.495{]} & 0.425 {[}0.365, 0.483{]} & 0.673 {[}0.622, 0.726{]} & 0.680 {[}0.624, 0.732{]} & \textbf{0.715 {[}0.664, 0.762{]}} \\
 & \begin{tabular}[c]{@{}c@{}}Respiratory\\ sounds\end{tabular} & 380 / 9851 (Y/N) & AP & 0.539 {[}0.489, 0.587{]} & 0.535 {[}0.489, 0.580{]} & 0.629 {[}0.583, 0.675{]} & 0.670 {[}0.624, 0.716{]} & \textbf{0.702 {[}0.652, 0.749{]}} \\
 & Sneeze & 61 / 10170 (Y/N) & AP & 0.361 {[}0.260, 0.471{]} & 0.448 {[}0.340, 0.559{]} & 0.570 {[}0.445, 0.685{]} & 0.650 {[}0.537, 0.746{]} & \textbf{0.912 {[}0.843, 0.964{]}} \\
 & Speech & 785 / 9446 (Y/N) & AP & 0.430 {[}0.397, 0.466{]} & 0.418 {[}0.384, 0.452{]} & 0.567 {[}0.533, 0.603{]} & 0.534 {[}0.498, 0.572{]} & \textbf{0.599 {[}0.568, 0.629{]}} \\
\midrule
FluSense & Breathing & 35 / 1624 (Y/N) & AP & 0.147 {[}0.088, 0.246{]} & 0.233 {[}0.135, 0.359{]} & 0.357 {[}0.238, 0.506{]} & 0.336 {[}0.236, 0.464{]} & \textbf{0.371 {[}0.238, 0.539{]}} \\
 & Cough & 430 / 721 (Y/N) & AP & 0.903 {[}0.881, 0.922{]} & 0.892 {[}0.870, 0.912{]} & 0.954 {[}0.941, 0.965{]} & \textbf{0.974 {[}0.966, 0.982{]}} & 0.963 {[}0.949, 0.974{]} \\
 & Gasp & 106 / 1780 (Y/N) & AP & 0.466 {[}0.384, 0.570{]} & 0.587 {[}0.499, 0.694{]} & 0.653 {[}0.568, 0.734{]} & 0.608 {[}0.518, 0.695{]} & \textbf{0.701 {[}0.606, 0.789{]}} \\
 & Sneeze & 160 / 1367 (Y/N) & AP & 0.648 {[}0.579, 0.714{]} & 0.661 {[}0.589, 0.727{]} & 0.810 {[}0.753, 0.860{]} & 0.788 {[}0.720, 0.848{]} & \textbf{0.825 {[}0.770, 0.877{]}} \\
 & Sniffle & 158 / 1413 (Y/N) & AP & 0.718 {[}0.654, 0.778{]} & 0.667 {[}0.599, 0.736{]} & 0.720 {[}0.648, 0.792{]} & \textbf{0.852 {[}0.799, 0.893{]}} & 0.841 {[}0.783, 0.889{]} \\
 & Speech & 483 / 593 (Y/N) & AP & 0.949 {[}0.937, 0.961{]} & 0.949 {[}0.937, 0.960{]} & \textbf{0.983 {[}0.976, 0.989{]}} & 0.972 {[}0.962, 0.981{]} & 0.973 {[}0.962, 0.983{]} \\
 & Throat-Clearing & 18 / 2057 (Y/N) & AP & 0.070 {[}0.022, 0.171{]} & 0.099 {[}0.029, 0.255{]} & 0.035 {[}0.026, 0.046{]} & \textbf{0.436 {[}0.243, 0.644{]}} & 0.169 {[}0.090, 0.314{]}

 \\
\bottomrule
\end{tabular}
}
% \vspace{-10pt}
\caption{Performance comparison on health acoustic event detection tasks on FSD50K and FluSense datasets. Due to high class imbalance, and following conventional reporting on those datasets, we report average precision (AP), together with bootstrapped 95\% confidence intervals. We also include mean average precision on the top row. Note that ``Respiratory sounds'' in FSD50K is the respiratory sound class excluding the other five FSD50K health acoustic event classes in the table.}
% \vspace{-10pt}
\label{tab2}
\end{table*}

\clearpage
When evaluated on cough inference tasks, HeAR performed better than the baselines across 10/14 tasks, including demographics, lifestyle, and COVID tasks. On TB and CXR tasks, its performance is comparable to the best performing model. A summary of results is listed in Table~\ref{tab3}. Appendix Table~\ref{tab_c1} compares the performance of all models across recording devices on CIDRZ. In particular, the performance of HeAR on the CXR most balanced tasks (lung opacities and unspecified abnormalities) exhibits the lowest variation (at most 1\% AUROC difference between the best and worst recording devices) and it reaches the highest performance for mid-tier and high-tier datasets, while TRILL and FRILL, while scoring highest AUROC on the low-tier dataset (Table~\ref{tab3}), also exhibit the highest variation, with up to 12\% drop between best and worst recording devices (Appendix Table~\ref{tab_c1}).

\begin{table*}[h]
\centering
\resizebox{1.0\textwidth}{!}{
\begin{tabular}{ccccccccc}
\toprule
\textbf{Dataset} & \textbf{Task} & \textbf{\begin{tabular}[c]{@{}c@{}}Summary statistics\\ for test split\\ (label)\end{tabular}} & \textbf{Metric} & \textbf{TRILL} & \textbf{FRILL} & \textbf{BigSSL-CAP12} & \textbf{HeAR} & \textbf{CLAP} \\
\midrule
CIDRZ (Pixel3a) & \begin{tabular}[c]{@{}c@{}}Focal / multi \\ focal lung opacities\end{tabular} & 61 / 204 (Y/N) & AUROC {[}DeLong 95\% CI{]} & \textbf{0.809 {[}0.747, 0.870{]}} & 0.800 {[}0.740, 0.860{]} & 0.747 {[}0.672, 0.821{]} & 0.794 {[}0.728, 0.861{]} & 0.760 {[}0.690, 0.830{]} \\
CIDRZ (Pixel3a) & Abnormal CXR & 64 / 201 (Y/N) & AUROC {[}DeLong 95\% CI{]} & \textbf{0.815 {[}0.757, 0.874{]}} & 0.778 {[}0.712, 0.844{]} & 0.739 {[}0.664, 0.814{]} & 0.763 {[}0.695, 0.830{]} & 0.734 {[}0.658, 0.810{]} \\
CIDRZ (Pixel3a) & Pleural effusion & 20 / 244 (Y/N) & AUROC {[}DeLong 95\% CI{]} & 0.683 {[}0.553, 0.812{]} & 0.688 {[}0.562, 0.813{]} & 0.684 {[}0.548, 0.819{]} & 0.610 {[}0.465, 0.755{]} & \textbf{0.748 {[}0.629, 0.866{]}} \\
CIDRZ (Pixel3a) & Tuberculosis & 24 / 240 (Y/N) & AUROC {[}DeLong 95\% CI{]} & 0.652 {[}0.520, 0.784{]} & 0.648 {[}0.523, 0.772{]} & 0.659 {[}0.533, 0.786{]} & 0.739 {[}0.636, 0.841{]} & \textbf{0.740 {[}0.627, 0.853{]}} \\
CIDRZ (Pixel3a) & Sex & 151 / 114 (F/M) & AUROC {[}DeLong 95\% CI{]} & 0.933 {[}0.901, 0.965{]} & 0.928 {[}0.894, 0.961{]} & 0.936 {[}0.909, 0.964{]} & \textbf{0.974 {[}0.958, 0.990{]}} & 0.907 {[}0.872, 0.942{]} \\
CoughVID & Sex & 1031 / 1924 (F/M) & AUROC {[}DeLong 95\% CI{]} & 0.850 {[}0.835, 0.866{]} & 0.848 {[}0.832, 0.863{]} & 0.872 {[}0.858, 0.887{]} & \textbf{0.897 {[}0.884, 0.910{]}} & 0.821 {[}0.805, 0.838{]} \\
Coswara & Sex & 174 / 478 (F/M) & AUROC {[}DeLong 95\% CI{]} & 0.920 {[}0.894, 0.947{]} & 0.917 {[}0.891, 0.942{]} & 0.937 {[}0.917, 0.956{]} & \textbf{0.979 {[}0.965, 0.993{]}} & 0.892 {[}0.862, 0.923{]} \\
CIDRZ (Pixel3a) & Smoking status & 65 / 198 (ever / never) & AUROC {[}DeLong 95\% CI{]} & 0.822 {[}0.762, 0.883{]} & 0.811 {[}0.747, 0.874{]} & 0.840 {[}0.786, 0.895{]} & \textbf{0.877 {[}0.833, 0.921{]}} & 0.808 {[}0.750, 0.866{]} \\
Coswara & Smoking status & 43 / 58 (current / never) & AUROC {[}DeLong 95\% CI{]} & 0.627 {[}0.518, 0.735{]} & 0.619 {[}0.509, 0.729{]} & 0.587 {[}0.476, 0.698{]} & \textbf{0.631 {[}0.523, 0.739{]}} & 0.619 {[}0.509, 0.729{]} \\
CoughVID & COVID status & 172 / 2237 (Y/N) & AUROC {[}DeLong 95\% CI{]} & 0.636 {[}0.565, 0.708{]} & 0.634 {[}0.562, 0.706{]} & 0.663 {[}0.596, 0.730{]} & \textbf{0.710 {[}0.647, 0.774{]}} & 0.618 {[}0.540, 0.696{]} \\
Coswara & COVID status & 63 / 470 (Y/N) & AUROC {[}DeLong 95\% CI{]} & 0.622 {[}0.581, 0.663{]} & 0.615 {[}0.574, 0.657{]} & 0.611 {[}0.570, 0.652{]} & \textbf{0.645 {[}0.603, 0.687{]}} & 0.624 {[}0.584, 0.665{]} \\
CIDRZ (Pixel3a) & BMI & \begin{tabular}[c]{@{}c@{}}$\mu$=22.9 kg/$m^2$\\ $\sigma$=6.01 kg/$m^2$\end{tabular} & \begin{tabular}[c]{@{}c@{}}Mean absolute error\\ {[}95\% bootstrapped CI{]}\end{tabular} & 3.861 {[}3.362, 4.458{]} & 3.860 {[}3.359, 4.465{]} & 3.875 {[}3.366, 4.452{]} & \textbf{3.818 {[}3.328, 4.397{]}} & 3.836 {[}3.337, 4.433{]} \\
CIDRZ (Pixel3a) & Age & \begin{tabular}[c]{@{}c@{}}$\mu$=35.6 yr\\ $\sigma$=13.1 yr\end{tabular} & \begin{tabular}[c]{@{}c@{}}Mean absolute error\\ {[}95\% bootstrapped CI{]}\end{tabular} & 10.590 {[}9.733, 11.509{]} & 10.959 {[}10.125, 11.855{]} & 10.009 {[}9.157, 10.944{]} & \textbf{9.316 {[}8.550, 10.123{]}} & 10.775 {[}9.934, 11.644{]} \\
Coswara & Age & \begin{tabular}[c]{@{}c@{}}$\mu$=32.7 yr \\ $\sigma$=12.3 yr\end{tabular} & \begin{tabular}[c]{@{}c@{}}Mean absolute error\\ {[}95\% bootstrapped CI{]}\end{tabular} & 9.994 {[}9.479, 10.570{]} & 10.013 {[}9.484, 10.590{]} & 9.665 {[}9.156, 10.247{]} & \textbf{8.742 {[}8.269, 9.277{]}} & 10.133 {[}9.592, 10.743{]}
 \\
\bottomrule
\end{tabular}
}
% \vspace{-10pt}
\caption{Performance comparison on cough inference tasks. We report AUROC (with DeLong 95\% confidence intervals) for binary classification tasks and mean absolute error (with bootstrapped confidence intervals) for regression tasks (age and BMI).}
% \vspace{-10pt}
\label{tab3}
\end{table*}

\clearpage
On SpiroSmart, HeAR performed better than our baselines on 4/5 lung function tasks and on sex classification. A summary of results is listed in Table~\ref{tab4}.

\begin{table*}[h]
\centering
\resizebox{1.0\textwidth}{!}{
\begin{tabular}{cccccccc}
\toprule
\textbf{Task} & \textbf{\begin{tabular}[c]{@{}c@{}}Summary statistics\\ for test split\end{tabular}} & \textbf{Metric} & \textbf{TRILL} & \textbf{FRILL} & \textbf{BigSSL-CAP12} & \textbf{HeAR} & \textbf{CLAP (48k)} \\
\midrule
FEV1 & \begin{tabular}[c]{@{}c@{}}108 patients\\ $\mu$=1.71L\\ $\sigma$=0.914L\end{tabular} & \begin{tabular}[c]{@{}c@{}}Mean absolute error\\ {[}95\% bootstrapped CI{]}\end{tabular} & 0.488 {[}0.408, 0.568{]} & 0.481 {[}0.404, 0.565{]} & 0.479 {[}0.404, 0.558{]} & \textbf{0.418 {[}0.351, 0.491{]}} & 0.518 {[}0.442, 0.604{]} \\
FVC & \begin{tabular}[c]{@{}c@{}}108 patients\\ $\mu$=2.29L\\ $\sigma$=0.915L\end{tabular} & \begin{tabular}[c]{@{}c@{}}Mean absolute error\\ {[}95\% bootstrapped CI{]}\end{tabular} & 0.559 {[}0.482, 0.641{]} & 0.548 {[}0.471, 0.631{]} & 0.536 {[}0.457, 0.614{]} & \textbf{0.476 {[}0.409, 0.547{]}} & 0.561 {[}0.483, 0.641{]} \\
FEV1/FVC & \begin{tabular}[c]{@{}c@{}}108 patients\\ $\mu$=0.717\\ $\sigma$=0.167\end{tabular} & \begin{tabular}[c]{@{}c@{}}Mean absolute error\\ {[}95\% bootstrapped CI{]}\end{tabular} & 0.087 {[}0.073, 0.102{]} & 0.090 {[}0.075, 0.105{]} & \textbf{0.083 {[}0.069, 0.098{]}} & \textbf{0.083 {[}0.070, 0.097{]}} & 0.086 {[}0.073, 0.103{]} \\
PEF & \begin{tabular}[c]{@{}c@{}}108 patients\\ $\mu$=4.78 L/s\\ $\sigma$=2.32 L/s\end{tabular} & \begin{tabular}[c]{@{}c@{}}Mean absolute error\\ {[}95\% bootstrapped CI{]}\end{tabular} & 1.199 {[}1.022, 1.375{]} & 1.318 {[}1.117, 1.498{]} & 1.319 {[}1.130, 1.506{]} & \textbf{1.147 {[}0.956, 1.343{]}} & 1.388 {[}1.192, 1.581{]} \\
FET & \begin{tabular}[c]{@{}c@{}}108 patients\\ $\mu$=6.68s\\ $\sigma$=2.47s\end{tabular} & \begin{tabular}[c]{@{}c@{}}Mean absolute error\\ {[}95\% bootstrapped CI{]}\end{tabular} & 1.540 {[}1.254, 1.937{]} & 1.541 {[}1.251, 1.932{]} & 1.452 {[}1.171, 1.819{]} & 1.508 {[}1.212, 1.878{]} & \textbf{1.371 {[}1.101, 1.703{]}} \\
Sex (male/female) & \begin{tabular}[c]{@{}c@{}}108 patients\\ 50 / 58 (M/F)\end{tabular} & AUROC {[}DeLong 95\% CI{]} & 0.914 {[}0.858, 0.970{]} & 0.914 {[}0.858, 0.970{]} & 0.931 {[}0.882, 0.980{]} & 0.934 {[}0.891, 0.978{]} & 0.878 {[}0.809, 0.947{]}
 \\
\bottomrule
\end{tabular}
}
% \vspace{-10pt}
\caption{Performance comparison on spirometry tasks. We report mean absolute error for regression tasks (with bootstrapped 95\% confidence intervals), and AUROC (with DeLong 95\% confidence intervals) for binary classification tasks (sex).}
% \vspace{-10pt}
\label{tab4}
\end{table*}

\clearpage
\newpage

% \vspace{-15pt}
\section{Discussion}
% \vspace{-5pt}
In this work, we develop and evaluate the HeAR system that integrates a health acoustic event detector and a generative learning-based audio encoder (MAE) to learn health acoustic representations. The audio encoder is trained on YT-NS, without the requirement of human or expert data curation. We then demonstrate the quality of acoustic representations learned from that system via health acoustic event detection tasks, cough inference tasks, and spirometry inference tasks. In particular, advances in classification of tuberculosis from cough sounds could help risk-stratify patients needing X-ray screening or further testing in environments where chest X-rays are scarce or unavailable. The potential to monitor lung function from smartphone audio recordings could help develop easy-to-use and ubiquitous COPD screening tools and doctors monitor more closely their patients' lung function evolution.

Self-supervised learning has achieved significant success in various domains, leveraging vast unlabeled data to train robust and generalizable encoders~\citep{radford2021learning,zhang2022bigssl,zhang2023google,devlin2018bert,raffel2020exploring,yu2022coca}. These encoders learn representations that have demonstrated increased robustness to distribution shift, improved transferability, and greater data efficiency~\citep{radford2021learning}. To the best of our knowledge, our work presents the first application of such large-scale self-supervised learning to health acoustic tasks. Our experiments reveal that increased pretraining data enhances downstream performance across diverse tasks (see Appendix D), consistent with prior findings~\citep{radford2021learning}. Notably, HeAR consistently achieves superior performance on diverse health-relevant tasks (inference of medical conditions and medically-relevant quantities from recordings of coughs or exhalations), as shown in Tables~\ref{tab1},\ref{tab3}, and \ref{tab4}. Since it is likely that users would use microphones not seen during training, robustness to recording devices is valuable. We show in Appendix C that HeAR’s performance on CIDRZ tasks remains stable across recording devices. In addition, when training linear probes on two devices and evaluating the probes on the third device (i.e., held-out device), HeAR performs better than other models (MRR=0.745 vs. second-best being CLAP with MRR=0.497), indicating potential for real-world applications. We hypothesize that the scale and diversity of recording devices in YT-NS contribute to making HeAR embeddings more device-agnostic. In addition, we find that HeAR is more data efficient than the baselines, sometimes reaching the same level of performance when trained on as little as 6.25\% of the amount of training data (see Appendix F). This is particularly relevant to instances where labeled training data is scarce, which is unfortunately commonplace in health research. Publicly available datasets are a highly valuable resource but remain scarce and typically have few participants, making it difficult to leverage modern deep learning techniques. Our approach addresses data scarcity by enabling models to achieve adequate performance with fewer training examples than traditional methods require.

On performance of these models, we should note that the reported performance for these tasks leverage linear probes and frozen embeddings, rather than fine-tuning the whole neural network~\citep{kohn2015s}. While this is common practice to evaluate models trained with SSL objectives and provides a fair comparison of different audio encoders, it may not yield the best performance for any given task. Performance could potentially be improved by including patient metadata as additional features of the linear probes or by fine tuning the full model.
For almost all tasks, confidence intervals are very wide due to our small datasets’ sizes and high class imbalance. This makes it hard to draw statistically significant conclusions, and further validation is required for such models to become part of clinically useful tools. Some datasets, like CIDRZ and SpiroSmart, are also specific samples of the population with high disease prevalence, so performance of the models trained on those datasets may not generalize to a healthier population, and further validation is required to estimate clinical usefulness of such tools for a general population. Importantly, though not unexpected since fundamental frequencies differ on average between sexes, the fact that the representations contain information about sex should be accounted for in future model development based on these representations~\citep{weng2024intentional}. For example, model performance should be examined stratified by sex, and any biases corrected as appropriate. It is also important to note that the performance reported on all tasks is not directly comparable to the literature because (1) prior dataset splits may not be described~\citep{bhattacharya2023coswara}, and (2) we use linear probing rather than full fine-tuning~\citep{kohn2015s}.

Other factors similarly affect generalization of these insights from these benchmark datasets. Evaluation on FSD50K and FluSense may not be representative of actual acoustic health events detection performance. FluSense audio clips are samples of Youtube videos, and we found that HeAR’s pretraining dataset YT-NS includes 172 of those videos (114 / 30 / 28 in train / validation / test), representing potentially 1394 clips (987 / 318 / 89 in train / validation / test). This could artificially inflate HeAR’s performance on FluSense. Similarly, BigSSL-CAP12 was trained on 900k hours of speech data from Youtube, which could also include samples from FluSense and inflate its measured performance on that dataset. On FSD50K, the timestamps of the acoustic events of interest are not available. Our coarse approximation of this information (the two-second around the loudest part of the clip) is likely to affect the labels and this is not accounted for in our training and evaluation procedures. Besides, this dataset was also used for training CLAP. These two reasons make it an imperfect benchmark.

The ad hoc preprocessing of FSD50K clips stems from the inability of most models to successfully generalize to inputs larger than what they have seen at training time. This is particularly acute in the case of HeAR, since its audio encoder is a transformer using 2-dimension fixed positional encoding trained on short two-second clips. That duration was chosen since it is typically longer than many of these health acoustic events. For tasks where the objective is to infer information about the participants (including all tasks from CIDRZ, Coswara, CoughVID, and SpiroSmart), only being able to process such short clips is sufficient. However, for detection tasks in longer audio clips like FSD50K’s and FluSense, this may not be sufficient (needle-in-a-haystack problem). The purpose of including those datasets in the evaluation procedure was to show that the same type of approach was useful for other types of sounds beyond cough alone, but specific adjustments would likely be necessary for reaching a more satisfying performance. Examples include modeling approaches not using transformers, for example with CNN and contrastive learning objectives, or using more appropriate positional encoding schemes~\citep{kazemnejad2024impact}.

Most models that we have experimented with are quite large and could not reasonably run on a smartphone. For reducing battery use, improving latency, preserving privacy, and making sure that this kind of technology can be used in the field with limited internet access, further research is needed to make sure that those models can run locally on-device, using techniques like distillation~\citep{hinton2015distilling} or quantization~\citep{gholami2022survey}.

We hope that our research can spur further research in the field of ML for health acoustic research. Individuals interested in getting access to use HeAR or the CIDRZ cough dataset can email \url{health_acoustic_representations@google.com} to be notified when available.

% \clearpage
% \newpage

\acks{We thank Yun Liu, Rory Pilgrim, Timo Kohlberger, Eduardo Fonseca, Aren Jansen, Dan Ellis, Ryan Ehrlich, Marc Wilson from Google Research, and Luyu Wang, Lucas Smaira, Eric Lau, Chung-Cheng Chiu, Basil Mustafa from Google DeepMind for their guidance, technical support and critical feedback. We also thank Solomon Chifwamba, Pauline Musumali, Kachimba Shamaoma, Seke Muzazu, Francesca Silwamba from the Centre for Infectious Disease Research in Zambia. We also appreciate the CoughVID and Project Coswara teams for making their respective datasets publicly available, and the Google Research team for software and hardware infrastructure support. CoughVID and Coswara are licensed under a \href{https://creativecommons.org/licenses/by/4.0/}{Creative Commons Attribution 4.0 International (CC BY 4.0) License} and follow the Disclaimer of Warranties and Limitation of Liability in the license.}

\bibliography{hear}
\clearpage

\newpage
\appendix
\section{Health Acoustic Event Detector}\label{apd:a}
The health acoustic event detector is trained on two publicly available datasets (FSD50K and Flusense) and a proprietary health acoustic dataset. FSD50K contains over 50K audio clips (over 100 hours) annotated using AudioSet ontology~\citep{fonseca2021fsd50k}, and FluSense is the subset of AudioSet dataset including sounds related to flu illnesses, which has seven labels with enough samples for running our cross-validation procedure: breathe, cough, gasp, sneeze, sniffle, speech, throat-clear, and ``etc'' (everything else)~\citep{al2020flusense}. The private dataset is collected from a variety of sources. The detector uses the audio clips with labels such as ``coughing'', ``sneezing'', and ``breathing'' for training.

The detector first converts and resamples the audio to mono channel 16 kHz sampling rate, then crops the audio into two-second log-mel spectrogram features with 48 frequency bins ranging from 125Hz to 7.5kHz with per-channel energy normalization (PCEN)~\citep{wang2017trainable}. These features are passed into a small convolutional neural network (CNN). The loss is balanced binary cross entropy, and the output of CNN is the logits for each prediction class. Detection yield for each event class in YouTube is listed in Table~\ref{tab_a1}. Two classes identified by the detector (sneezing and snoring) were not used for filtering YouTube samples because they were not deemed reliable enough.

\setcounter{table}{0}
\renewcommand{\thetable}{A\arabic{table}}

\begin{table}[h]
\footnotesize
\centering
% \resizebox{\textwidth}{!}{
\begin{tabular}{cc}
\toprule
\textbf{Sound Type} & \textbf{\begin{tabular}[c]{@{}c@{}}Yield\\ (number of two-second audio clips)\end{tabular}} \\
\midrule
Coughing & 50,414,000 \\
Baby coughing & 1,411,000 \\
Breathing & 31,534,560 \\
Throat clearing & 4,095,000 \\
Laughing & 102,826,000 \\
Speaking & 123,024,000 \\
\bottomrule
\end{tabular}
% }
\caption{Detection yield for each health acoustic event from three billions YouTube clips.}
\label{tab_a1}
\end{table}

\section{Evaluation Datasets}\label{apd:b}
The details of five evaluation datasets are listed in Table~\ref{tab_b1}. Note that some datasets include more than one recording per participant. When this happens, predictions obtained on recordings from a given individual are averaged. Metrics in all tables (in particular, Tables~\ref{tab1},\ref{tab2},\ref{tab3},\ref{tab_c1}) are computed on a per-participant level (and not per-recording).

For FluSense, there may be several acoustic events occurring in each clip. When that happens, there will be several labeled crops extracted from a given clip. Predictions in that case are not aggregated at the clip level, hence the identical counts in both columns for that dataset. 

\setcounter{table}{0}
\renewcommand{\thetable}{B\arabic{table}}

\begin{table*}[h]
\footnotesize
\centering
\resizebox{\textwidth}{!}{
\begin{tabular}{cccccc}
\toprule
\textbf{Dataset} & \textbf{Tasks} & \textbf{\begin{tabular}[c]{@{}c@{}}Number of recordings\\ used for training \\ linear probes\end{tabular}} & \textbf{\begin{tabular}[c]{@{}c@{}}Number of recordings\\ (participants) used \\ for validation\end{tabular}} & \textbf{\begin{tabular}[c]{@{}c@{}}Number of recordings\\ (participants) for\\ final evaluation\end{tabular}} & \textbf{Reference} \\
\midrule
FSD50K & \begin{tabular}[c]{@{}c@{}}Health acoustic event \\ (6 tasks)\end{tabular} & 32652 & 8313 (8313) & 10231 (10231) &~\cite{fonseca2021fsd50k} \\
\midrule
Flusense & \begin{tabular}[c]{@{}c@{}}Health acoustic event \\ (7 tasks)\end{tabular} & 7535 & 1779 (1779) & 2360 (2360) &~\cite{al2020flusense} \\
\midrule
CoughVID & \begin{tabular}[c]{@{}c@{}}COVID, sex \\ (2 tasks)\end{tabular} & 44249 & 15083 (4095) & 15123 (2955) &~\cite{orlandic2021coughvid} \\
\midrule
Coswara & \begin{tabular}[c]{@{}c@{}}COVID, sex, smoking status, age \\ (4 tasks)\end{tabular} & 10230 & 4285 (531) & 5846 (652) &~\cite{bhattacharya2023coswara} \\
\midrule
CIDRZ & \begin{tabular}[c]{@{}c@{}}TB, sex, smoking status, \\ age, BMI, 3 CXR findings \\ (8 tasks, 3 different devices)\end{tabular} & 8210 & \begin{tabular}[c]{@{}c@{}}663 (86) (low-tier)\\ 854 (84) (mid-tier)\\ 857 (86) (high-tier)\end{tabular} & \begin{tabular}[c]{@{}c@{}}2107 (265) (low-tier)\\ 2460 (265)  (mid-tier)\\ 2546 (272) (high-tier)\end{tabular} & \begin{tabular}[c]{@{}c@{}}Dataset collected for this study. \\ Details in the CIDRZ section below.\end{tabular} \\
\midrule
SpiroSmart & \begin{tabular}[c]{@{}c@{}}FEV1, FVC, FEV1/FVC ratio, \\ PEF, FET, sex (6 tasks)\end{tabular} & 13239 & 696 (544) & 772 (108) &~\cite{garrison2018spiro} \\
\bottomrule
\end{tabular}
}
\caption{Evaluation datasets statistics.}
\label{tab_b1}
\end{table*}

\paragraph{CIDRZ TB Dataset Description}\label{apd:b1}
The CIDRZ TB dataset is collected by the Centre for Infectious Disease Research in Zambia. The study was approved by the University of Zambia Biomedical Ethics Committee, and all participants provided written informed consent prior to enrollment in the study. Adults who had symptoms suggestive of TB, were identified as close contacts of TB patients, or were newly diagnosed with HIV were recruited at three clinical sites (Chawama, Chainda-South, and Kanyama) in Zambia (trial NCT05139940). Audio recordings of cough sounds were obtained from 599 consented patients. To ensure robustness across different microphones, the sounds were recorded by four devices: Zoom H2N microphone (high quality audio recorder), Samsung GalaxyA22 (high-tier phone), Samsung GalaxyA12 (mid-tier phone), and Pixel3a (low-tier phone). In this work, we focused on recordings from the three phone microphones. The audio clips are recorded by the \href{https://gitlab.com/axet/android-audio-recorder}{Android application Audio Recorder} and encoded in the wav file format with 24-bit PCM, sampling rate of 192 kHz, and in stereo channels under a quiet environment. Cohort details are listed in Table~\ref{tab_b2}.

To collect cough sounds, the participant was asked to remove his/her mask and generate four cough events (three single coughs and one sequence of multiple coughs). There is a 10-15 seconds gap between cough events to enable a return to ``baseline'' before the next cough.

The CXR and the corresponding CXR finding annotations of the CIDRZ TB Dataset have been collected along with the cough sound collection.

\begin{table*}[h]
\footnotesize
\centering
\resizebox{\textwidth}{!}{
\begin{tabular}{cccccc}
\toprule
                        &                      & \textbf{All}                  & \textbf{Train}                & \textbf{Tune}                 & \textbf{Test}                 \\
\midrule
Sample size (\%)        &                      & 599 (100.0)          & 229 (38.2)           & 89 (14.9)            & 281 (46.9)           \\
\midrule
Site (\%)               & Chawama              & 281 (46.9)           & 0                    & 0                    & 281 (100.0)          \\
                        & Chainda-South        & 34 (5.7)             & 26 (11.4)            & 8 (9.0)              & 0                    \\
                        & Kanyama              & 284 (47.4)           & 203 (88.6)           & 81 (91.0)            & 0                    \\
\midrule
Female (\%)             &                      & 297 (49.6)           & 99 (43.2)            & 41 (46.1)            & 157 (55.9)           \\
\midrule
Age {[}IQR{]}           &                      & 35.0 {[}27.0,45.0{]} & 36.0 {[}28.0,46.0{]} & 37.0 {[}29.0,44.0{]} & 33.0 {[}25.0,45.0{]} \\
\midrule
BMI {[}IQR{]}           &                      & 21.0 {[}19.0,24.0{]} & 20.0 {[}18.0,24.0{]} & 20.0 {[}18.0,23.0{]} & 21.0 {[}19.0,25.0{]} \\
\midrule
Positive TB (\%)        &                      & 92 (15.4)            & 46 (20.1)            & 18 (20.2)            & 28 (10.0)            \\
\midrule
Positive HIV (\%)       &                      & 217 (36.2)           & 85 (37.1)            & 36 (40.4)            & 96 (34.2)            \\
\midrule
Cough duration (\%)     & 1 - 2 weeks          & 227 (37.9)           & 78 (34.1)            & 28 (31.5)            & 121 (43.1)           \\
                        & 3 - 4 weeks          & 24 (4.0)             & 5 (2.2)              & 3 (3.4)              & 16 (5.7)             \\
                        & \textless 1 week     & 121 (20.2)           & 52 (22.7)            & 17 (19.1)            & 52 (18.5)            \\
                        & \textgreater 4 weeks & 155 (25.9)           & 68 (29.7)            & 28 (31.5)            & 59 (21.0)            \\
\midrule
Productive cough (\%)   &                      & 455 (76.0)           & 181 (79.0)           & 68 (76.4)            & 206 (73.3)           \\
\midrule
Hemoptysis (\%)         &                      & 65 (10.9)            & 25 (10.9)            & 12 (13.5)            & 28 (10.0)            \\
\midrule
Chest pain (\%)         &                      & 375 (62.6)           & 149 (65.1)           & 63 (70.8)            & 163 (58.0)           \\
\midrule
Short of breath (\%)    &                      & 204 (34.1)           & 84 (36.7)            & 35 (39.3)            & 85 (30.2)            \\
\midrule
Fever (\%)              &                      & 217 (36.2)           & 86 (37.6)            & 38 (42.7)            & 93 (33.1)            \\
\midrule
Night sweat (\%)        &                      & 245 (40.9)           & 96 (41.9)            & 36 (40.4)            & 113 (40.2)           \\
\midrule
Weight loss (\%)        &                      & 374 (62.4)           & 157 (68.6)           & 49 (55.1)            & 168 (59.8)           \\
\midrule
Previous TB (\%)        & 0                    & 497 (83.0)           & 187 (81.7)           & 70 (78.7)            & 240 (85.4)           \\
                        & 1                    & 92 (15.4)            & 38 (16.6)            & 17 (19.1)            & 37 (13.2)            \\
                        & 2                    & 8 (1.3)              & 3 (1.3)              & 2 (2.2)              & 3 (1.1)              \\
                        & 3                    & 2 (0.3)              & 1 (0.4)              & 0                    & 1 (0.4)              \\
\midrule
Tobacco use (\%)        & Current              & 127 (21.2)           & 55 (24.0)            & 22 (24.7)            & 50 (17.8)            \\
                        & Stopped              & 48 (8.0)             & 16 (7.0)             & 11 (12.4)            & 21 (7.5)             \\
                        & Never                & 422 (70.5)           & 158 (69.0)           & 56 (62.9)            & 208 (74.0)           \\
\midrule
Cigarettes per day (\%) & No                   & 472 (78.8)           & 174 (76.0)           & 67 (75.3)            & 231 (82.2)           \\
                        & 1 - 10               & 93 (15.5)            & 37 (16.2)            & 16 (18.0)            & 40 (14.2)            \\
                        & 11 - 20              & 20 (3.3)             & 12 (5.2)             & 3 (3.4)              & 5 (1.8)              \\
                        & \textgreater 20      & 14 (2.3)             & 6 (2.6)              & 3 (3.4)              & 5 (1.8)             \\
\bottomrule
\end{tabular}
}
\caption{CIDRZ cohort descriptive statistics per split. Metadata field varies; the following table reports data where available.}
\label{tab_b2}
\end{table*}

\clearpage
\newpage
\section{CIDRZ TB Dataset Performance Per Recording Device Type}\label{apd:c}
The performance of BigSSL-CAP12, HeAR, and CLAP is stable across recording devices on most tasks, while TRILL's and FRILL's vary significantly between low-tier and high-tier (Table~\ref{tab_c1}). MRR for those tasks are 0.381, 0.274, 0.386, 0.786, and 0.456 for TRILL, FRILL, BigSSL-CAP12, HeAR, and CLAP, respectively.

\setcounter{table}{0}
\renewcommand{\thetable}{C\arabic{table}}

\begin{table*}[h]
\footnotesize
\centering
\resizebox{\textwidth}{!}{
\begin{tabular}{ccccccc}
\toprule
\textbf{\begin{tabular}[c]{@{}c@{}}Evaluation\\ Recording Device\end{tabular}} & \textbf{Task} & \textbf{TRILL} & \textbf{FRILL} & \textbf{BigSSL-12} & \textbf{HeAR} & \textbf{CLAP} \\
\midrule
Pixel3a & \begin{tabular}[c]{@{}c@{}}Focal / multi focal \\ lung opacities\end{tabular} & \textbf{0.809 {[}0.747, 0.870{]}} & 0.800 {[}0.740, 0.860{]} & 0.747 {[}0.672, 0.821{]} & 0.794 {[}0.728, 0.861{]} & 0.760 {[}0.690, 0.830{]} \\
GalaxyA12 & \begin{tabular}[c]{@{}c@{}}Focal / multi focal \\ lung opacities\end{tabular} & 0.743 {[}0.671, 0.814{]} & 0.728 {[}0.655, 0.802{]} & 0.760 {[}0.688, 0.831{]} & \textbf{0.802 {[}0.734, 0.870{]}} & 0.746 {[}0.673, 0.820{]} \\
GalaxyA22 & \begin{tabular}[c]{@{}c@{}}Focal / multi focal \\ lung opacities\end{tabular} & 0.719 {[}0.646, 0.792{]} & 0.686 {[}0.611, 0.762{]} & 0.746 {[}0.672, 0.820{]} & \textbf{0.806 {[}0.744, 0.867{]}} & 0.758 {[}0.686, 0.830{]} \\
Pixel3a & Abnormal CXR & \textbf{0.815 {[}0.757, 0.874{]}} & 0.778 {[}0.712, 0.844{]} & 0.739 {[}0.664, 0.814{]} & 0.763 {[}0.695, 0.830{]} & 0.734 {[}0.658, 0.810{]} \\
GalaxyA12 & Abnormal CXR & 0.730 {[}0.655, 0.806{]} & 0.734 {[}0.662, 0.807{]} & 0.734 {[}0.662, 0.807{]} & \textbf{0.763 {[}0.691, 0.836{]}} & 0.743 {[}0.673, 0.813{]} \\
GalaxyA22 & Abnormal CXR & 0.729 {[}0.657, 0.801{]} & 0.725 {[}0.653, 0.797{]} & 0.732 {[}0.658, 0.806{]} & \textbf{0.768 {[}0.701, 0.836{]}} & 0.746 {[}0.674, 0.818{]} \\
Pixel3a & Pleural effusion & 0.683 {[}0.553, 0.812{]} & 0.688 {[}0.562, 0.813{]} & 0.684 {[}0.548, 0.819{]} & 0.610 {[}0.465, 0.755{]} & \textbf{0.748 {[}0.629, 0.866{]}} \\
GalaxyA12 & Pleural effusion & \textbf{0.673 {[}0.564, 0.781{]}} & 0.632 {[}0.481, 0.784{]} & 0.635 {[}0.499, 0.771{]} & 0.625 {[}0.476, 0.774{]} & 0.567 {[}0.408, 0.725{]} \\
GalaxyA22 & Pleural effusion & 0.637 {[}0.521, 0.754{]} & 0.567 {[}0.433, 0.701{]} & 0.713 {[}0.587, 0.838{]} & 0.634 {[}0.493, 0.776{]} & \textbf{0.730 {[}0.614, 0.846{]}} \\
Pixel3a & Sex & 0.933 {[}0.901, 0.965{]} & 0.928 {[}0.894, 0.961{]} & 0.936 {[}0.909, 0.964{]} & \textbf{0.974 {[}0.958, 0.990{]}} & 0.907 {[}0.872, 0.942{]} \\
GalaxyA12 & Sex & 0.944 {[}0.917, 0.971{]} & 0.941 {[}0.912, 0.969{]} & 0.931 {[}0.899, 0.963{]} & \textbf{0.981 {[}0.969, 0.993{]}} & 0.912 {[}0.877, 0.947{]} \\
GalaxyA22 & Sex & 0.951 {[}0.925, 0.976{]} & 0.950 {[}0.924, 0.976{]} & 0.939 {[}0.910, 0.967{]} & \textbf{0.981 {[}0.966, 0.996{]}} & 0.900 {[}0.864, 0.936{]} \\
Pixel3a & Tuberculosis & 0.652 {[}0.520, 0.784{]} & 0.648 {[}0.523, 0.772{]} & 0.659 {[}0.533, 0.786{]} & 0.739 {[}0.636, 0.841{]} & \textbf{0.740 {[}0.627, 0.853{]}} \\
GalaxyA12 & Tuberculosis & 0.637 {[}0.527, 0.747{]} & 0.631 {[}0.503, 0.758{]} & 0.680 {[}0.571, 0.790{]} & 0.720 {[}0.617, 0.824{]} & \textbf{0.745 {[}0.641, 0.848{]}} \\
GalaxyA22 & Tuberculosis & 0.675 {[}0.574, 0.775{]} & 0.581 {[}0.457, 0.704{]} & 0.687 {[}0.584, 0.791{]} & 0.734 {[}0.639, 0.829{]} & \textbf{0.794 {[}0.700, 0.889{]}} \\
Pixel3a & Smoking status & 0.822 {[}0.762, 0.883{]} & 0.811 {[}0.747, 0.874{]} & 0.840 {[}0.786, 0.895{]} & \textbf{0.877 {[}0.833, 0.921{]}} & 0.808 {[}0.750, 0.866{]} \\
GalaxyA12 & Smoking status & 0.831 {[}0.772, 0.890{]} & 0.830 {[}0.771, 0.888{]} & 0.836 {[}0.781, 0.891{]} & \textbf{0.862 {[}0.817, 0.907{]}} & 0.797 {[}0.723, 0.871{]} \\
GalaxyA22 & Smoking status & 0.812 {[}0.753, 0.871{]} & 0.809 {[}0.749, 0.868{]} & 0.835 {[}0.777, 0.892{]} & \textbf{0.861 {[}0.811, 0.910{]}} & 0.787 {[}0.721, 0.853{]} \\
Pixel3a & Age & 10.590 {[}9.733, 11.509{]} & 10.959 {[}10.125, 11.855{]} & 10.009 {[}9.157, 10.944{]} & \textbf{9.316 {[}8.550, 10.123{]}} & 10.775 {[}9.934, 11.644{]} \\
GalaxyA12 & Age & 10.819 {[}9.962, 11.680{]} & 10.898 {[}10.000, 11.740{]} & 10.175 {[}9.366, 10.928{]} & \textbf{9.280 {[}8.505, 10.049{]}} & 10.610 {[}9.770, 11.408{]} \\
GalaxyA22 & Age & 11.285 {[}10.429, 12.137{]} & 11.332 {[}10.458, 12.216{]} & 10.006 {[}9.203, 10.840{]} & \textbf{9.554 {[}8.787, 10.375{]}} & 10.695 {[}9.881, 11.581{]} \\
Pixel3a & BMI & 3.861 {[}3.362, 4.458{]} & 3.860 {[}3.359, 4.465{]} & 3.875 {[}3.366, 4.452{]} & \textbf{3.818 {[}3.328, 4.397{]}} & 3.836 {[}3.337, 4.433{]} \\
GalaxyA12 & BMI & 3.853 {[}3.376, 4.448{]} & 3.845 {[}3.358, 4.441{]} & 3.781 {[}3.286, 4.369{]} & \textbf{3.719 {[}3.202, 4.270{]}} & 3.786 {[}3.288, 4.378{]} \\
GalaxyA22 & BMI & 3.792 {[}3.324, 4.345{]} & 3.793 {[}3.315, 4.342{]} & 3.747 {[}3.295, 4.263{]} & \textbf{3.695 {[}3.247, 4.206{]}} & 3.711 {[}3.250, 4.252{]} \\
\bottomrule
\end{tabular}
}
\caption{Performance comparison on cough inference tasks of CIDRZ TB Dataset per recording device type.}
\label{tab_c1}
\end{table*}

\clearpage
\newpage
\section{Effect of Scaling Training Data Size}\label{apd:d}
We also find that scaling up the training data (YT-NS) used for training the HeAR audio encoder helps improve the linear probing performance across different downstream tasks (Figure~\ref{fig_d1}).

\setcounter{figure}{0}
\renewcommand{\thefigure}{D\arabic{figure}}

\begin{figure*}[h]
\centering
\includegraphics[width=1\linewidth]{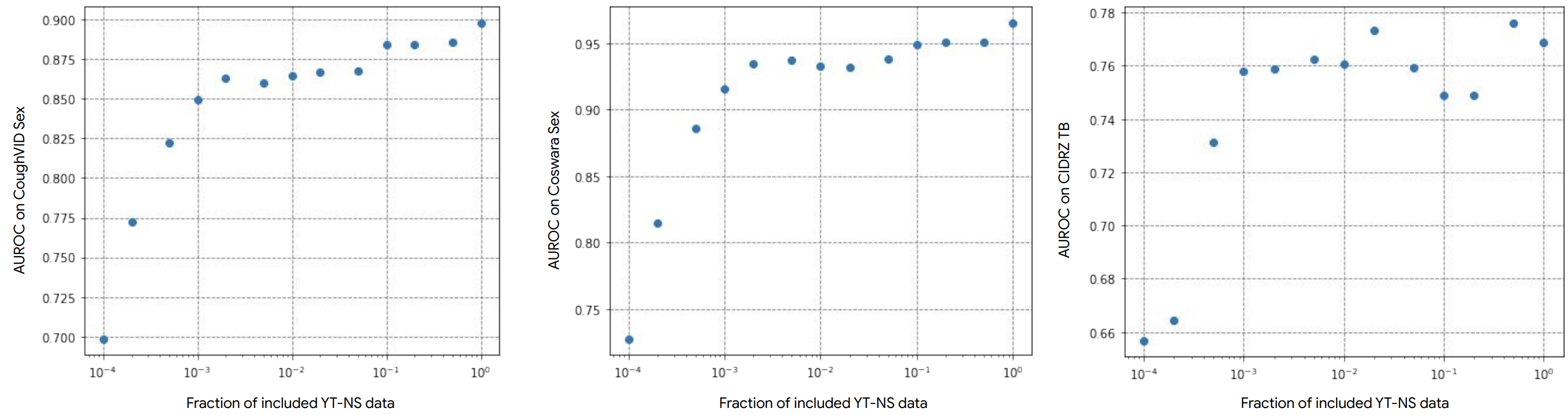} 
\caption{Scaling effect of increasing the YT-NS data size for training HeAR. We use CoughVID and Coswara sex classification, and CIDRZ tuberculosis prediction tasks as examples.}
\label{fig_d1}
\end{figure*}

\section{Generalization to Unseen Devices}\label{apd:e}
Two devices are used for training the linear probes, and evaluation is done using recordings from the remaining device (i.e., out-of-distribution (OOD) device). MRR for those tasks are 0.382, 0.303, 0.358, 0.743, and 0.497 for TRILL, FRILL, BigSSL-CAP12, HeAR, and CLAP, respectively.

For all tasks, HeAR performance remains stable across all OOD devices and is consistently among the highest ranked. Other models like TRILL and FRILL have unstable performance, while CLAP and BigSSL-CAP12 are more stable but typically worse (Table~\ref{tab_e1}).

\setcounter{table}{0}
\renewcommand{\thetable}{E\arabic{table}}

\begin{table}[h]
\resizebox{\textwidth}{!}{
\begin{tabular}{ccccccc}
\toprule
\textbf{OOD device} & \textbf{Task} & \textbf{TRILL} & \textbf{FRILL} & \textbf{BigSSL-CAP12} & \textbf{HeAR} & \textbf{CLAP} \\
\midrule
Pixel3a & \begin{tabular}[c]{@{}c@{}}Focal / multi \\ focal lung opacities\end{tabular} & \textbf{0.802 {[}0.741, 0.864{]}} & 0.759 {[}0.692, 0.827{]} & 0.744 {[}0.669, 0.819{]} & 0.789 {[}0.722, 0.856{]} & 0.769 {[}0.699, 0.840{]} \\
Pixel3a & Abnormal CXR & \textbf{0.788 {[}0.723, 0.852{]}} & 0.733 {[}0.662, 0.805{]} & 0.734 {[}0.661, 0.808{]} & 0.765 {[}0.699, 0.831{]} & 0.759 {[}0.688, 0.830{]} \\
Pixel3a & Pleural effusion & 0.672 {[}0.527, 0.817{]} & \textbf{0.720 {[}0.599, 0.841{]}} & 0.603 {[}0.462, 0.744{]} & 0.577 {[}0.430, 0.724{]} & 0.657 {[}0.528, 0.786{]} \\
Pixel3a & Sex & 0.926 {[}0.893, 0.958{]} & 0.920 {[}0.885, 0.954{]} & 0.937 {[}0.909, 0.964{]} & \textbf{0.973 {[}0.956, 0.989{]}} & 0.886 {[}0.845, 0.926{]} \\
Pixel3a & Tuberculosis & 0.669 {[}0.538, 0.801{]} & 0.651 {[}0.522, 0.779{]} & 0.696 {[}0.575, 0.816{]} & \textbf{0.748 {[}0.646, 0.849{]}} & 0.659 {[}0.530, 0.788{]} \\
Pixel3a & Smoking status & 0.816 {[}0.756, 0.877{]} & 0.805 {[}0.742, 0.868{]} & 0.835 {[}0.780, 0.889{]} & \textbf{0.878 {[}0.835, 0.921{]}} & 0.758 {[}0.688, 0.828{]} \\
GalaxyA12 & \begin{tabular}[c]{@{}c@{}}Focal / multi \\ focal lung opacities\end{tabular} & 0.706 {[}0.631, 0.782{]} & 0.730 {[}0.656, 0.803{]} & 0.751 {[}0.678, 0.824{]} & \textbf{0.800 {[}0.731, 0.869{]}} & 0.771 {[}0.703, 0.839{]} \\
GalaxyA12 & Abnormal CXR & 0.709 {[}0.633, 0.785{]} & 0.679 {[}0.602, 0.756{]} & 0.726 {[}0.653, 0.800{]} & 0.761 {[}0.688, 0.835{]} & \textbf{0.768 {[}0.703, 0.832{]}} \\
GalaxyA12 & Pleural effusion & 0.591 {[}0.448, 0.734{]} & 0.642 {[}0.533, 0.751{]} & 0.598 {[}0.456, 0.741{]} & 0.624 {[}0.470, 0.778{]} & \textbf{0.674 {[}0.533, 0.815{]}} \\
GalaxyA12 & Sex & 0.950 {[}0.924, 0.975{]} & 0.948 {[}0.922, 0.974{]} & 0.932 {[}0.901, 0.964{]} & \textbf{0.981 {[}0.969, 0.993{]}} & 0.904 {[}0.866, 0.941{]} \\
GalaxyA12 & Tuberculosis & 0.667 {[}0.568, 0.766{]} & 0.625 {[}0.503, 0.747{]} & 0.699 {[}0.595, 0.804{]} & \textbf{0.720 {[}0.621, 0.819{]}} & 0.715 {[}0.603, 0.826{]} \\
GalaxyA12 & Smoking status & 0.829 {[}0.771, 0.887{]} & 0.834 {[}0.777, 0.891{]} & 0.839 {[}0.784, 0.894{]} & \textbf{0.871 {[}0.827, 0.915{]}} & 0.759 {[}0.682, 0.836{]} \\
GalaxyA22 & \begin{tabular}[c]{@{}c@{}}Focal / multi \\ focal lung opacities\end{tabular} & 0.719 {[}0.645, 0.793{]} & 0.671 {[}0.593, 0.749{]} & 0.738 {[}0.664, 0.812{]} & \textbf{0.798 {[}0.735, 0.861{]}} & 0.771 {[}0.702, 0.840{]} \\
GalaxyA22 & Abnormal CXR & 0.711 {[}0.637, 0.786{]} & 0.732 {[}0.661, 0.803{]} & 0.728 {[}0.655, 0.802{]} & 0.772 {[}0.705, 0.839{]} & \textbf{0.774 {[}0.707, 0.841{]}} \\
GalaxyA22 & Pleural effusion & 0.620 {[}0.491, 0.749{]} & 0.548 {[}0.415, 0.682{]} & 0.729 {[}0.607, 0.852{]} & 0.679 {[}0.550, 0.808{]} & \textbf{0.757 {[}0.641, 0.872{]}} \\
GalaxyA22 & Sex & 0.951 {[}0.925, 0.977{]} & 0.937 {[}0.907, 0.967{]} & 0.944 {[}0.916, 0.971{]} & \textbf{0.982 {[}0.967, 0.997{]}} & 0.905 {[}0.869, 0.940{]} \\
GalaxyA22 & Tuberculosis & 0.561 {[}0.444, 0.678{]} & 0.592 {[}0.466, 0.717{]} & 0.633 {[}0.525, 0.741{]} & 0.727 {[}0.632, 0.822{]} & \textbf{0.746 {[}0.642, 0.849{]}} \\
GalaxyA22 & Smoking status & 0.816 {[}0.759, 0.873{]} & 0.798 {[}0.735, 0.861{]} & 0.833 {[}0.774, 0.892{]} & \textbf{0.856 {[}0.806, 0.906{]}} & 0.771 {[}0.704, 0.838{]} \\
\bottomrule
\end{tabular}
}
\caption{Performance on cough inference tasks of CIDRZ TB Dataset with the unseen device generalization setup.}
\label{tab_e1}
\end{table}

\section{Data efficiency}\label{apd:f}
For all cough and spirometry tasks, we train linear probes with 6.25\%, 12.5\%, 25\%, 50\%, and 100\% of the data, for all models. All models use the exact same subsampled datasets for training. All probes are evaluated on 100\% of the test split. For classification tasks, we make sure to subsample each label in the same way. This procedure allows us to compare how different encoders fare in different data regimes. We find that HeAR is the most data efficient model, as evidenced by its consistently higher rank across all data regimes and all tasks (Table~\ref{tab_f1}, Figure~\ref{fig_f1},\ref{ref_f2}).

\setcounter{table}{0}
\renewcommand{\thetable}{F\arabic{table}}

\begin{table}[h]
\begin{tabular}{cccccc}
\toprule
\textbf{Training data available (\%)} & \textbf{TRILL} & \textbf{FRILL} & \textbf{BigSSL-CAP12} & \textbf{HeAR} & \textbf{CLAP (48k)} \\
\midrule
6.25 & 0.409 & 0.392 & 0.358 & \textbf{0.789} & 0.331 \\
12.5 & 0.436 & 0.310 & 0.390 & \textbf{0.840} & 0.304 \\
25 & 0.321 & 0.308 & 0.470 & \textbf{0.843} & 0.342 \\
50 & 0.392 & 0.358 & 0.384 & \textbf{0.790} & 0.358 \\
100 & 0.418 & 0.341 & 0.359 & \textbf{0.796} & 0.359 \\
\bottomrule
\end{tabular}
\caption{Mean reciprocal rank (MRR) across all cough (Coswara, CoughVID, CIDRZ) and spirometry (SpiroSmart) inference tasks, for varying amounts of available training data.}
\label{tab_f1}
\end{table}

\setcounter{figure}{0}
\renewcommand{\thefigure}{F\arabic{figure}}

\begin{figure*}[h]
\centering
\includegraphics[width=1\linewidth]{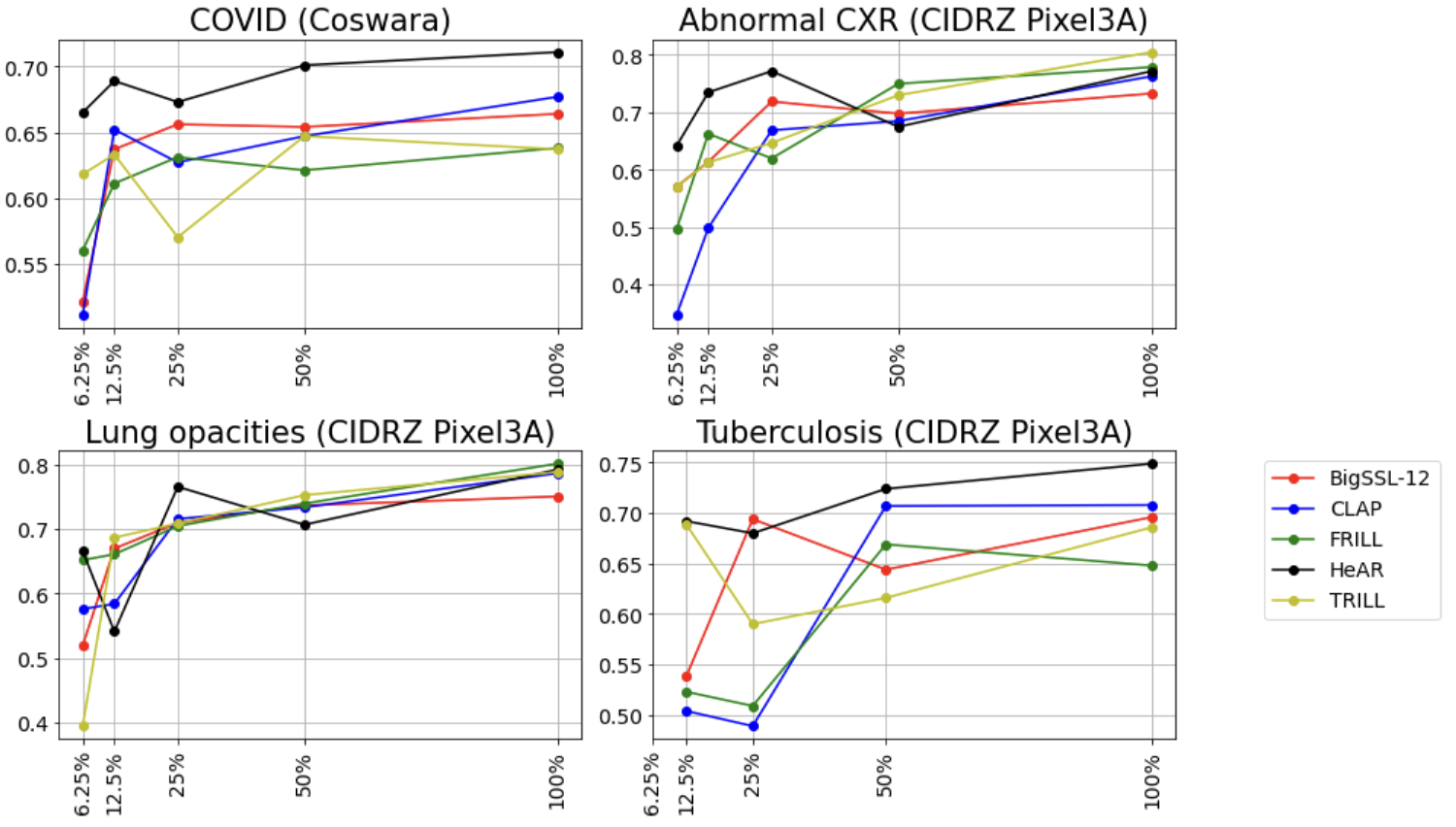} 
\caption{AUROC of all models for varying amounts of training data, for a subset of cough inference tasks.}
\label{fig_f1}
\end{figure*}

\begin{figure*}[h]
\centering
\includegraphics[width=1\linewidth]{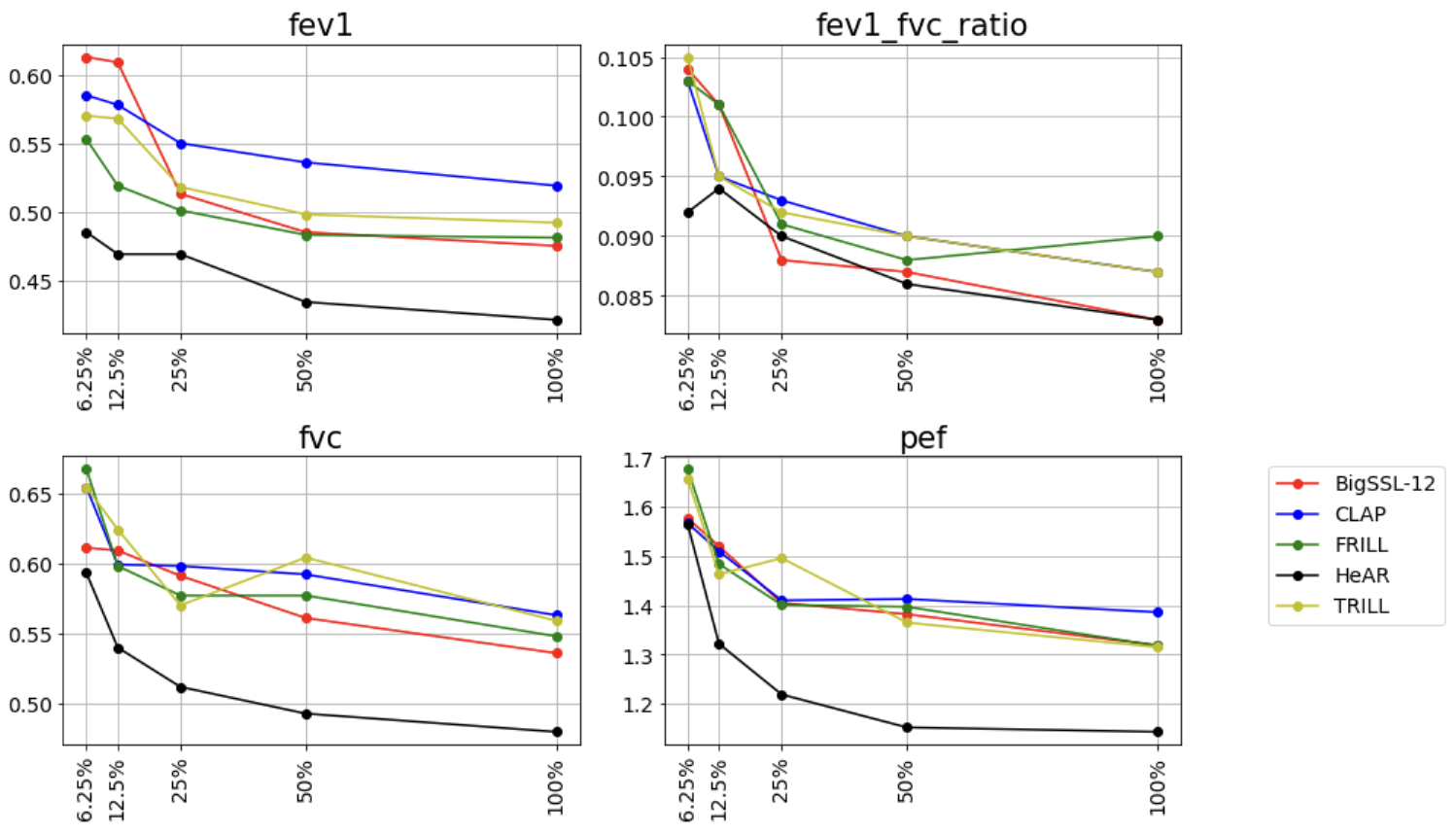} 
\caption{Mean absolute error of all models for varying amounts of training data, for a subset of spirometry inference tasks.}
\label{ref_f2}
\end{figure*}

\end{document}